\documentclass[sigconf]{acmart}

\AtBeginDocument{%
  }

\setcopyright{none}
\renewcommand\footnotetextcopyrightpermission[1]{}

\usepackage{multirow}
\usepackage{makecell}
\usepackage{mathtools}
\usepackage{pifont}
\usepackage{algorithm}
\usepackage{algorithmic}
\usepackage{enumitem}
\usepackage{colortbl}

\theoremstyle{remark}

\newcommand{\ours}{DynaForcing}

\newcommand{\etal}{\textit{et al.}}

\newcommand{\ul}[1]{\underline{#1}}

\definecolor{isabelline}{RGB}{244,241,234}
\definecolor{lightblue}{RGB}{218,232,252}

\newcommand{\tablestyle}[2]{%
  \setlength{\tabcolsep}{#1}%
  \renewcommand{\arraystretch}{#2}%
}

\newcommand{\shline}{%
  \noalign{\global\setlength{\arrayrulewidth}{0.2pt}}%
  \hline
  \noalign{\global\setlength{\arrayrulewidth}{0.4pt}}%
}

\begin{document}

\title{DynaForcing: Overcoming Dynamic Collapse in Self-Forcing Distillation for Streaming Avatar Generation}

\author{Yubo Huang}
\email{snake1124@mail.ustc.edu.cn}
\orcid{0009-0005-8870-0216}
\affiliation{%
  \institution{University of Science and Technology of China}
  \city{Hefei}
  \country{China}
}

\author{Sirui Zhao}
\email{sirui@mail.ustc.edu.cn}
\orcid{0000-0001-8103-0321}
\affiliation{%
  \institution{University of Science and Technology of China}
  \city{Hefei}
  \country{China}
}

\author{Xinchen Yao}
\email{xinchenyao@smail.nju.edu.cn}
\orcid{0009-0008-7045-0337}
\affiliation{%
  \institution{Nanjing University}
  \city{Nanjing}
  \country{China}
}

\author{Zhengye Zhang}
\email{zhengyezhang@mail.ustc.edu.cn}
\orcid{0009-0008-1558-827X}
\affiliation{%
  \institution{University of Science and Technology of China}
  \city{Hefei}
  \country{China}
}

\author{Jinyang Huang}
\email{hjy@hfut.edu.cn}
\orcid{0000-0001-5483-2812}
\affiliation{%
  \institution{Hefei University of Technology}
  \city{Hefei}
  \country{China}
}

\author{Fengqi Cui}
\email{fengqi\_cui@mail.ustc.edu.cn}
\orcid{0000-0002-0454-940X}
\affiliation{%
  \institution{University of Science and Technology of China}
  \city{Hefei}
  \country{China}
}

\author{Shiwei Wu}
\email{davidwu16@sz.tsinghua.edu.cn}
\orcid{0000-0002-3206-6827}
\affiliation{%
  \institution{Tsinghua University}
  \city{Shenzhen}
  \country{China}
}

\author{Enhong Chen}
\authornote{Corresponding author.}
\email{cheneh@ustc.edu.cn}
\orcid{0000-0002-4835-4102}
\affiliation{%
  \institution{University of Science and Technology of China}
  \city{Hefei}
  \country{China}
}

\renewcommand{\shortauthors}{Yubo Huang et al.}

\begin{abstract}

Audio-driven avatar generation requires realistic lip-sync, expressive motion, and real-time streaming. Recent work achieves the latter via self-forcing with Distribution Matching Distillation (DMD), but this paradigm suffers from a critical failure that has not been systematically characterized: \textbf{dynamic collapse}, where the student model converges to a near-static optimum with high perceptual quality but severely suppressed temporal dynamics. We trace this to two causes: the reverse KL objective in DMD, which biases toward low-motion modes, and unanchored self-conditioning, which creates a feedback loop that amplifies collapse. This is especially harmful for avatars, where even subtle motion loss breaks lip-sync and expression.
To address this, we propose \textbf{\ours{}}, a training framework with three complementary strategies applied at different levels. Specifically, \emph{Hybrid Forcing} anchors rollouts to ground-truth dynamics at the data level to break the feedback loop. \emph{Dynamics-Aware Reward Regularization} introduces explicit motion rewards via the RL interpretation of DMD to counteract the reverse KL bias at the loss level. \emph{Reference Perturbation} perturbs reference images to decouple identity from static details, forcing the model to rely on audio for motion at the conditioning level. We further introduce computation graph pruning and gradient replay, reducing the GPU footprint of self-forcing by over an order of magnitude. Experiments show that \ours{} recovers dynamics to teacher-comparable levels (Dyn-Deg: 0.31$\to$0.73, Sync-C: 7.03$\to$7.68) while improving visual quality, resolving the quality--dynamics trade-off throughout training without early stopping.

\end{abstract}



\begin{teaserfigure}
  \centering
  \includegraphics[width=0.80\linewidth]{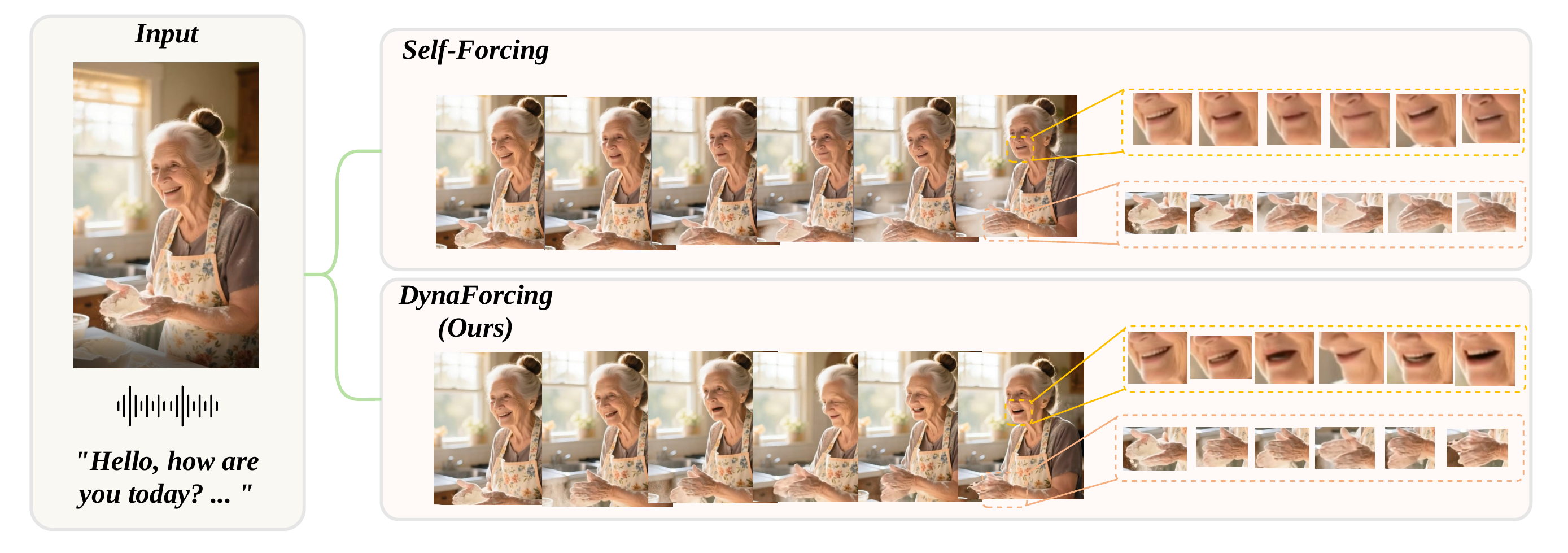}
  \caption{Dynamic collapse in self-forcing distillation for streaming avatar generation. Given the same reference image and driving audio (left), the self-forcing baseline (top) produces visually appealing but temporally frozen outputs---nearly identical lip shapes across 10 seconds of speech (see mouth crops). \ours{} (bottom) restores faithful audio-driven dynamics with natural expression variation and accurate lip articulation, at 45.2\,FPS real-time streaming.}
  \label{fig:teaser}
\end{teaserfigure}

\maketitle

\pagestyle{plain}
\thispagestyle{plain}

\section{Introduction}
\label{sec:intro}

Audio-driven avatar generation aims to synthesize photorealistic talking-head video whose motion faithfully follows an input audio stream, serving as  a cornerstone technology for interactive digital communication~\cite{prajwal2020lip,tian2024emo,gan2025omniavatar}. Deploying such systems in real-world scenarios, such as video conferencing and virtual assistants, requires satisfying two simultaneous demands: \emph{real-time streaming} at interactive frame rates, and \emph{temporally rich dynamics} including precise lip articulation, natural expressions, and responsive head motion. Recent large-scale video diffusion models~\cite{wan2025wan,videoworldsimulators2024} have significantly advanced visual fidelity, and self-forcing~\cite{huang2025self,cui2025self} has emerged as a leading distillation paradigm, leveraging Distribution Matching Distillation (DMD)~\cite{yin2024one} to convert these models into causal, few-step streaming generators that enable real-time, infinite-length avatar generation at 14B-parameter scale~\cite{huang2025liveavatar,soulxflashtalk2025}.

\begin{figure*}[t]
  \centering
  \includegraphics[width=0.96\linewidth]{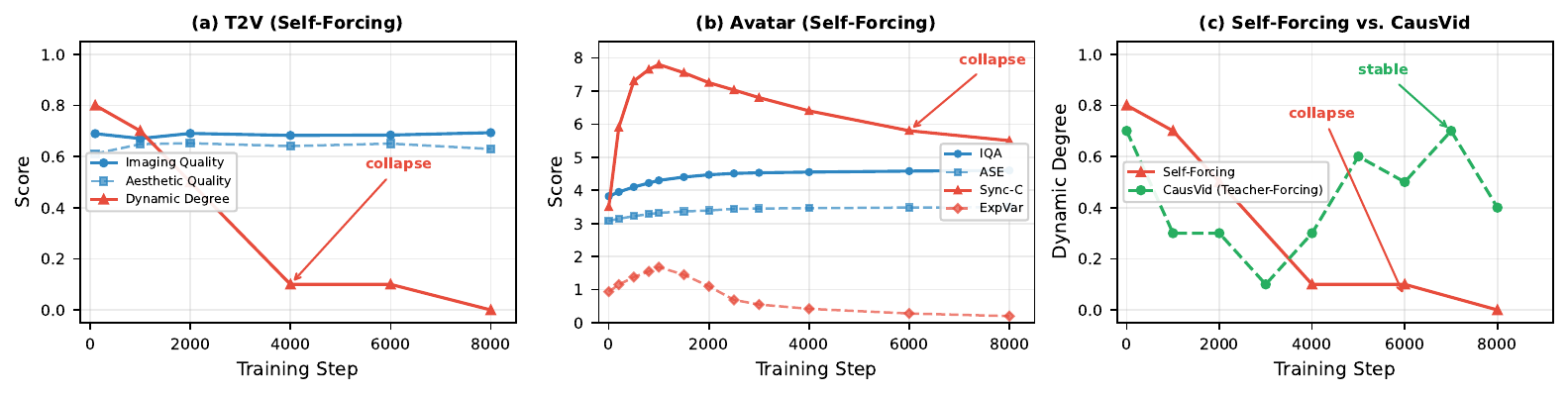}
  \caption{Dynamic collapse in self-forcing distillation. (a)~Text-to-video (Self-Forcing~\cite{huang2025self}): VBench~\cite{huang2023vbench} \emph{dynamic degree} drops from 0.80 to 0.00 while visual quality improves. (b)~Audio-driven avatar (LiveAvatar~\cite{huang2025liveavatar}): Sync-C and ExpVar degrade while IQA and ASE improve. (c)~Comparing self-forcing with CausVid~\cite{yin2025slow} (GT-anchored): CausVid's dynamic degree fluctuates around 0.43 without collapsing, confirming that unanchored self-conditioning, not DMD distillation itself, drives the collapse.}
  \label{fig:dynamic_collapse}
\end{figure*}

Yet beneath these impressive capabilities lies a failure mode that, while noted in recent work~\cite{liu2026streaming,rewardforcing2025}, has not been systematically characterized. We find that self-forcing training systematically drives the student model toward what we term \textbf{dynamic collapse} (Figure~\ref{fig:teaser}), where generated videos exhibit high perceptual quality scores but near-zero temporal dynamics: mouths barely move, expressions freeze, and the rich motion present in the teacher model vanishes. Avatar generation is particularly susceptible because the task requires \emph{fine-grained} temporal dynamics (precise lip shapes for each phoneme, micro-expressions, head motion responsive to prosody), yet the static-scene prior (fixed background, stable identity) means that a near-motionless output already satisfies most visual quality criteria, leaving little gradient pressure toward dynamics. While DMD's diversity--quality tradeoff is well-documented for images~\cite{yin2024one,yin2024improved}, its impact on video is qualitatively different: collapsing temporal diversity does not merely reduce ``variety'' but fundamentally breaks the core requirement of faithful audio-visual correspondence.

We trace this failure to two contributing mechanisms. First, DMD's reverse KL objective $D_{\mathrm{KL}}(p_\theta \| p_{\text{data}})$ is mode-seeking~\cite{murphy2012machine}: it penalizes the student for placing mass outside the teacher's support but imposes no cost for ignoring modes. Second, self-forcing performs \emph{unanchored self-conditioning}: unlike GT-anchored approaches (e.g., CausVid~\cite{yin2025slow}) that process all blocks in parallel with ground-truth latents as input, self-forcing builds the entire KV cache from the student's own outputs with no ground-truth anchoring whatsoever. This creates a compounding feedback loop: when the student begins producing low-dynamics content, all subsequent blocks condition on this self-generated static history, reinforcing the collapse across the entire sequence. We verify this empirically: under identical training horizons, CausVid's GT-anchored conditioning maintains stable dynamics while self-forcing collapses to near-zero, as demonstrated in Figure~\ref{fig:dynamic_collapse}c.

To address dynamic collapse, we propose \textbf{\ours{}}, a training framework with three complementary strategies operating at different levels of the training pipeline:

\begin{enumerate}[leftmargin=*,nosep]
\item \textbf{Hybrid Forcing} (\S\ref{sec:hybrid_forcing}), a \emph{data-level} strategy that probabilistically replaces self-forcing rollout starting points with noised ground-truth latents. This provides a ground-truth anchor that counteracts mode collapse in the distillation objective.

\item \textbf{Dynamics-Aware Reward Regularization} (\S\ref{sec:reward_reg}), a \emph{loss-level} strategy that leverages the RL interpretation of DMD~\cite{luo2024diff} to introduce explicit dynamics rewards (lip-sync accuracy and expression variance) as auxiliary training signals that directly penalize static outputs.

\item \textbf{Reference Perturbation} (\S\ref{sec:ref_diversity}), a complementary \emph{conditioning-level} strategy that perturbs reference images to decouple identity from extraneous visual details (pose, background, lighting), preventing the model from taking a shortcut of copying the reference frame instead of generating audio-driven motion.
\end{enumerate}

Additionally, we present \textbf{efficient self-forcing training} (\S\ref{sec:efficiency}) via computation graph pruning and gradient replay, reducing GPU requirements by over $10\times$ while preserving quality.

In summary, the main contributions of this work are as follows:
\begin{itemize}[leftmargin=*,nosep]
\item We formalize and systematically characterize \emph{dynamic collapse} in self-forcing distillation, tracing it to reverse-KL mode-seeking and the compounding effect of unanchored self-conditioning.
\item We propose \ours{}, combining three strategies at different levels (data, loss, conditioning) that jointly address dynamic collapse while preserving visual quality.
\item We introduce computation graph pruning and gradient replay for self-forcing, achieving $\sim$10$\times$ GPU-hour reduction.
\item Experiments on audio-driven avatar generation demonstrate substantial improvements in motion naturalness and audio-visual faithfulness on both short- and long-form benchmarks.
\end{itemize}

\section{Related Work}
\label{sec:related}

\noindent\textbf{Audio-Driven Avatar Generation.}
Audio-driven talking-head synthesis has evolved from GAN-based methods~\cite{prajwal2020lip,zhang2023sadtalker} to diffusion-based approaches~\cite{tian2024emo,gan2025omniavatar,cui2025hallo3,gao2025wans2vaudiodrivencinematicvideo,huang2026bindyouravatar,xu2026emotionally} that leverage large-scale video diffusion models for dramatically improved fidelity. More recently, several works have achieved \emph{real-time streaming} avatar generation by combining diffusion models with autoregressive distillation~\cite{huang2025liveavatar,soulxflashtalk2025,wang2026flowactr1,sun2025streamavatar,ki2026avatarforcing}, demonstrating interactive-rate generation at scale. However, these streaming methods consistently exhibit \emph{significantly degraded motion precision and naturalness} compared to their multi-step teachers, even though the underlying DMD distillation has been shown to produce students that match or exceed teacher quality on static image metrics~\cite{yin2024one,yin2024improved}. For avatars, this motion loss is particularly damaging: lip-sync accuracy, expression diversity, and gesture naturalness are the core requirements, and any dynamics degradation directly undermines the task objective.

\noindent\textbf{Streaming Video Distillation.}
Distilling bidirectional multi-step diffusion models into few-step causal streaming generators has been pursued through progressive distillation~\cite{salimans2022progressive}, consistency models~\cite{song2023consistency}, and DMD~\cite{yin2024one,yin2024improved}. CausVid~\cite{yin2025slow} introduced GT-anchored autoregressive distillation, and Self-Forcing~\cite{huang2025self,cui2025self} further removed ground-truth anchoring by conditioning entirely on the student's own outputs, better aligning train-time and test-time distributions. Subsequent works have refined this paradigm: Causal Forcing~\cite{zhu2026causalforcing} bridges the architectural gap via AR-teacher ODE initialization, Rolling Forcing~\cite{liu2025rolling} suppresses error accumulation through joint multi-frame denoising, and LongLive~\cite{yang2025longlive} extends stable generation to longer horizons. The field has largely converged on DMD-based self-forcing as the dominant paradigm for streaming video generation. Yet self-forcing introduces an inherent \emph{motion degradation} problem: the combination of reverse-KL mode-seeking and unanchored self-conditioning creates a feedback loop that systematically suppresses temporal dynamics. This has been independently observed by DiagDistill~\cite{liu2026streaming}, which proposes diagonal forcing with implicit flow modeling to mitigate motion amplitude attenuation, and Reward Forcing~\cite{rewardforcing2025}, which introduces reward signals to encourage dynamics. Nevertheless, the problem remains largely unsolved, especially for tasks like avatar generation that demand fine-grained, audio-synchronized motion rather than coarse temporal coherence.

\begin{figure*}[t]
  \centering
  \includegraphics[width=0.80\linewidth]{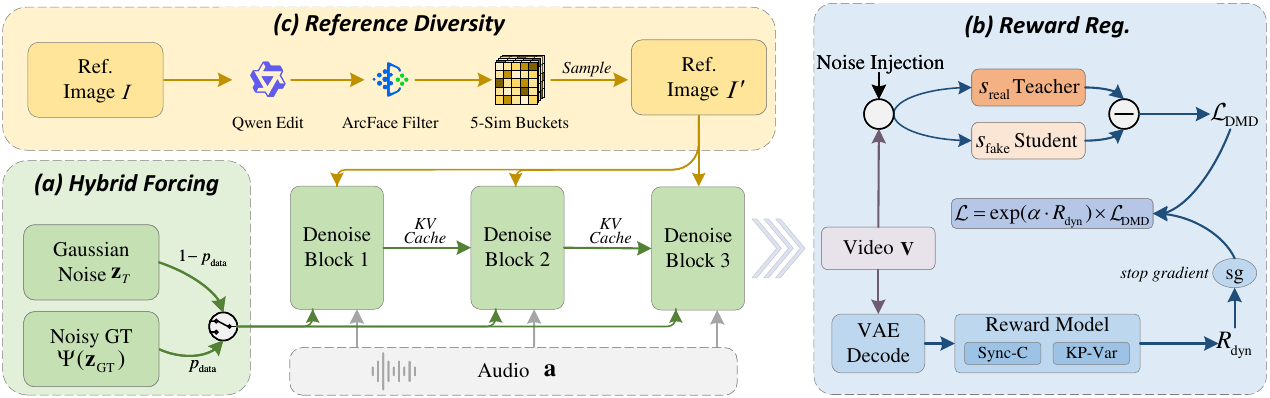}
  \caption{Overview of \ours{}. Three strategies intervene at different levels of the self-forcing training pipeline: (a) Hybrid Forcing at the input level, (b) dynamics-aware reward regularization at the loss level, and (c) reference perturbation at the conditioning level.}
  \label{fig:method_overview}
\end{figure*}

\section{Preliminaries}
\label{sec:prelim}

\subsection{Self-Forcing for Streaming Video Diffusion}
\label{sec:prelim_sf}

Self-forcing~\cite{huang2025self} enables streaming video generation by combining autoregressive rollout with few-step diffusion sampling. The video is generated block-by-block, where each block $B^i$ contains $L$ consecutive frame latents. At each rollout step, the student model $G_\theta$ denoises the current block conditioned on a rolling KV cache of previous blocks:
\begin{equation}
\hat{B}^i_0 = G_\theta(B^i_T, \underbrace{B^{(i-w):({i-1})}_{t}}_{\text{KV cache}}, I, a^i),
\label{eq:rollout}
\end{equation}
where $B^i_T \sim \mathcal{N}(0, \mathbf{I})$ is the initial noise, $w$ is the cache window size, $I$ is the reference (sink) frame, $a^i$ is the audio embedding, and $t$ denotes the shared noise level of the KV cache (timestep-forcing). The student produces a multi-step denoising trajectory $B^i_T \to B^i_{t_1} \to \cdots \to \hat{B}^i_0$, and the generated clean latent $\hat{B}^i_0$ is used to construct the KV cache for subsequent blocks.

\subsection{Distribution Matching Distillation as RL}
\label{sec:prelim_dmd_rl}

Distribution Matching Distillation (DMD)~\cite{yin2024one} trains the student by minimizing the reverse KL divergence between the student's induced distribution $p_\theta$ and the teacher's data distribution $p_{\text{data}}$, averaged over noise levels $t$:
\begin{equation}
\mathcal{L}_{\text{DMD}} = \mathbb{E}_t \left[ D_{\mathrm{KL}} \big( p_{\theta,t} \,\|\, p_{\text{data},t} \big) \right].
\label{eq:dmd_kl}
\end{equation}
The gradient of this objective takes the form:
\begin{equation}
\nabla_\theta \mathcal{L}_{\text{DMD}} = -\mathbb{E}_{t, \mathbf{z}} \left[ \big( s_{\text{real}}(\mathbf{x}_t, t) - s_{\text{fake}}(\mathbf{x}_t, t) \big)^\top \frac{\partial G_\theta(\mathbf{z})}{\partial \theta} \right],
\label{eq:dmd_grad}
\end{equation}
where $s_{\text{real}}$ is the teacher (pretrained) score and $s_{\text{fake}}$ is a learned score on student-generated data, and $\mathbf{x}_t = \Psi(\hat{\mathbf{x}}, t)$ is the noise-perturbed student output.

Crucially, Luo~\etal{}~\cite{luo2024diff} frame DMD within an RL/RLHF objective: the score difference in Eq.~\ref{eq:dmd_grad} serves as an Integral KL regularization toward the teacher, and additional reward signals can be incorporated by augmenting the objective with an explicit reward term. Building on this, Lu~\etal{}~\cite{rewardforcing2025} propose \emph{reward-weighted} distillation, where an external reward $R(\mathbf{x})$ exponentially reweights the DMD gradient:
\begin{equation}
\nabla_\theta \mathcal{L}_{\text{Re-DMD}} = -\mathbb{E}_{t, \mathbf{z}} \left[ \exp\!\big(\alpha \cdot \overline{R}\big) \cdot \big( s_{\text{real}} - s_{\text{fake}} \big)^\top \frac{\partial G_\theta(\mathbf{z})}{\partial \theta} \right],
\label{eq:dmd_rl}
\end{equation}
where $\overline{R} = \mathrm{sg}(R)$ is the stopped-gradient reward and $\alpha$ controls the reward strength. This formulation amplifies the DMD gradient for high-reward samples while suppressing it for low-reward ones, effectively steering the student toward reward-rich modes without requiring backpropagation through the reward model.

This reward-weighted framework is central to our approach: by defining a dynamics-specific reward, we can bias the self-forcing training loop away from degenerate static modes (\S\ref{sec:reward_reg}).

\section{Method: DynaForcing}
\label{sec:method}

We first analyze the dynamic collapse phenomenon (\S\ref{sec:dynamic_collapse}), then present our three strategies: Hybrid Forcing (\S\ref{sec:hybrid_forcing}), Dynamics-Aware Reward Regularization (\S\ref{sec:reward_reg}), and Reference Perturbation (\S\ref{sec:ref_diversity}). Finally, we describe efficient training (\S\ref{sec:efficiency}).

\subsection{Understanding Dynamic Collapse}
\label{sec:dynamic_collapse}

We observe that self-forcing training can severely suppress temporal dynamics despite producing visually appealing outputs. We term this phenomenon \textbf{dynamic collapse} and analyze its causes.

\noindent\textbf{Empirical evidence.}
Figure~\ref{fig:dynamic_collapse} demonstrates that dynamic collapse is not task-specific but a general pathology of self-forcing distillation.\footnote{Reproduction details for Figure~\ref{fig:dynamic_collapse} are provided in the supplementary material.} On text-to-video generation (Figure~\ref{fig:dynamic_collapse}a), VBench~\cite{huang2023vbench} \emph{dynamic degree} monotonically decreases from 0.80 to 0.00 over 8K training steps, while \emph{imaging quality} and \emph{aesthetic quality} steadily improve. On audio-driven avatar generation (Figure~\ref{fig:dynamic_collapse}b), the same pattern emerges: visual quality (IQA, ASE) improves steadily while dynamics metrics (Sync-C, ExpVar) peak early and then decline sharply, eventually reaching severely suppressed levels. In both cases, the generated videos look sharp and artifact-free but exhibit minimal temporal variation.

\noindent\textbf{Analysis: Mode-seeking reverse KL.}
DMD minimizes $D_{\mathrm{KL}}(p_\theta \| p_{\text{data}})$, which is mode-seeking~\cite{murphy2012machine}: it penalizes mass outside the teacher's support but incurs no cost for ignoring modes. In the video domain, static content is an attractive mode because the DMD score difference $s_{\text{real}} - s_{\text{fake}}$ provides limited gradient signal for dynamics, as the teacher's score function primarily captures distributional realism, not temporal richness. As the student concentrates on a narrow, low-motion subset of the teacher's distribution, the fake score $s_{\text{fake}}$ quickly matches $s_{\text{real}}$ in this region, eliminating the gradient signal needed to explore other modes.

\noindent\textbf{Analysis: Unanchored self-conditioning.}
Self-forcing builds the entire KV cache from the student's own outputs, with no ground-truth anchoring during the rollout. Once the student begins producing low-motion blocks, all subsequent blocks condition on this static history, creating a compounding feedback loop that reinforces collapse across the sequence. The key insight is not autoregression per se, but the \emph{absence of any ground-truth signal in the conditioning path}. Figure~\ref{fig:dynamic_collapse}c validates this analysis: under identical evaluation, CausVid~\cite{yin2025slow}, which conditions each block on noisy ground-truth frames, maintains stable dynamics (mean 0.43) while self-forcing collapses to 0.

\noindent\textbf{Analysis: Why avatar generation is particularly affected.}
While dynamic collapse can occur in general video distillation, audio-driven avatar generation is particularly susceptible because: (i) the task requires \emph{fine-grained} temporal dynamics (precise lip shapes for each phoneme, micro-expressions), making any dynamics reduction immediately noticeable; and (ii) the static-scene prior (fixed background, stable identity) means that a zero-motion output already satisfies most of the visual quality requirements, leaving little gradient pressure toward dynamics.

\subsection{Hybrid Forcing: Data-Anchored Rollout}
\label{sec:hybrid_forcing}

To break the feedback loop of dynamic collapse, we propose \emph{Hybrid Forcing}, which probabilistically injects ground-truth dynamics into the self-forcing rollout at the \emph{starting point} level, providing an anchor that counteracts mode collapse under unanchored self-conditioning.

At each training step, we sample a mode for the entire rollout (i.e., the decision is per-sample, applied uniformly to all blocks). Denoting the full sequence starting point as $\mathbf{z} = [B^1_{\text{start}}, \ldots, B^K_{\text{start}}]$, each block's starting point is:
\begin{equation}
B^i_{\text{start}} =
\begin{cases}
B^i_T \sim \mathcal{N}(0, \mathbf{I}), & \text{w.p. } 1 - p_{\text{data}}, \\
\Psi(B^i_{\text{GT}}, t_{\text{start}}), & \text{w.p. } p_{\text{data}},
\end{cases}
\label{eq:hybrid}
\end{equation}
where $B^i_{\text{GT}}$ is the ground-truth video latent, $\Psi(\cdot, t)$ is the forward noise scheduler, and $t_{\text{start}} \sim \{t_0, \ldots, t_{K-1}\}$ is sampled from the $K$ noise levels in the few-step denoising schedule. In data-anchored mode, the student starts from a noised ground-truth rather than pure noise, preserving temporal dynamics (motion patterns, lip movements) as structural information that the student must refine.

The DMD loss is then computed on the student's denoised output regardless of the starting mode. When starting from data-anchored points, the loss gradient points toward dynamically-rich outputs, providing an escape direction from the zero-motion basin. The KV cache always uses the student's own representations (noisy, per timestep-forcing), maintaining train-test alignment.

\noindent\textbf{Why data-anchored starting points help.}
A noised ground-truth latent $\Psi(x_{\text{GT}}, t) = \alpha_t x_{\text{GT}} + \sigma_t \varepsilon$ retains recoverable structure (motion cues, temporal dynamics, spatial layout) as long as the signal-to-noise ratio $\alpha_t/\sigma_t$ is not too small. A pure Gaussian start $z \sim \mathcal{N}(0, \mathbf{I})$ carries none of this information, forcing the student to reconstruct everything from its imperfect learned prior. We formalize this intuition in the supplementary material, showing that data-anchored starts yield a tighter Wasserstein bound on the denoised output.

\noindent\textbf{Starting-step probability redistribution.}
Starting from $t_{\text{start}}$ means the model only traverses the denoising sub-trajectory $t_{\text{start}} \to \cdots \to t_0$ (Eq.~\ref{eq:hybrid}), so noisier levels receive less training coverage. Backward simulation~\cite{yin2024improved}, which self-forcing inherits, addresses a similar coverage issue for pure-noise starts. To compensate in the data-anchored case, we redistribute the sampling probability of $t_{\text{start}}$: since $t_0$ already operates at high SNR where strong dynamic priors are preserved and little denoising correction is needed, we concentrate mass on early steps, which retain the most dynamic structure from the ground truth while still requiring non-trivial denoising.

\subsection{Dynamics-Aware Reward Regularization}
\label{sec:reward_reg}

\noindent\textbf{Framework.}
As discussed in \S\ref{sec:prelim_dmd_rl}, DMD can be viewed through an RL lens where the score difference serves as an implicit reward~\cite{luo2024diff}. However, this implicit reward can be fully satisfied by mode collapse (concentrating on high-likelihood static content), since it measures distributional realism, not temporal richness. We introduce an explicit dynamics reward $R_{\text{dyn}}$ as an additional return signal that directly penalizes static outputs. Following the reward-weighted distillation framework~\cite{luo2024diff,rewardforcing2025}, we integrate $R_{\text{dyn}}$ by exponentially reweighting the DMD gradient:
\begin{equation}
\resizebox{0.91\linewidth}{!}{$\displaystyle
\nabla_\theta \mathcal{L}_{\text{DynaForcing}} = -\mathbb{E}_{t, \mathbf{z}} \!\left[ \exp\!\big(\alpha \cdot \mathrm{sg}(R_{\text{dyn}})\big) \cdot \big( s_{\text{real}} - s_{\text{fake}} \big)^\top \frac{\partial G_\theta(\mathbf{z})}{\partial \theta} \right]
$},
\label{eq:dynaforcing_grad}
\end{equation}
where $\mathrm{sg}(\cdot)$ denotes the stop-gradient operator and $\alpha$ controls the reward strength. This amplifies the DMD gradient for samples with rich dynamics and suppresses it for static outputs, approximating optimization toward a reward-tilted target $\tilde{p}(\mathbf{x}) \propto p_{\text{teacher}}(\mathbf{x}) \exp(\alpha R_{\text{dyn}}(\mathbf{x}))$, analogous to reward-weighted regression in offline RL~\cite{peters2008reinforcement}.

\noindent\textbf{Reward implementation.}
Let $\hat{\mathbf{V}}\!=\![\hat{B}^1_0,\ldots,\hat{B}^K_0]$ denote the student-rolled video.
We define the dynamics reward as:
\begin{equation}
R_{\text{dyn}}(\hat{\mathbf{V}}, \mathbf{a}) = \lambda_{\text{sync}} \cdot \widetilde{\text{Sync-C}}(\hat{\mathbf{V}}, \mathbf{a}) + \lambda_{\text{exp}} \cdot \widetilde{\text{ExpVar}}(\hat{\mathbf{V}}),
\label{eq:reward}
\end{equation}
where $\widetilde{\text{Sync-C}}$ is the normalized lip-sync confidence from a pretrained SyncNet~\cite{syn} and $\widetilde{\text{ExpVar}}$ is the normalized standard deviation of 3DMM expression coefficients across frames~\cite{zhu2025infp}. Both are computed on decoded video frames and normalized to $[0, 1]$ via running statistics. $R_{\text{dyn}}$ is treated as a constant during backpropagation (stop-gradient), so only the score terms and generator Jacobian carry gradient flow, making this fully compatible with standard DMD training.

\subsection{Reference Perturbation}
\label{sec:ref_diversity}

We identify that standard training practice induces a complementary source of dynamics suppression at the conditioning level: the reference image leaks far more information than identity alone.

\noindent\textbf{The copy-dominance problem.}
In avatar generation, the reference image $I$ is typically sampled from the same video clip as the target. Since reference and target share scene, background, viewpoint, and often similar pose, the model learns a shortcut: \emph{copy the reference frame} with minimal modification, rather than generating genuine audio-driven dynamics. The reference is meant to supply only identity, but in practice it leaks far more, suppressing the need for temporal generation.

\noindent\textbf{Perturbing references to decouple identity.}
To decouple identity from extraneous visual details, we perturb reference images so that they retain identity but vary in other attributes. Specifically, we use an off-the-shelf image editing model~\cite{wu2025qwenimagetechnicalreport} to generate perturbed reference images from the training set. Given a reference frame $I$, we apply semantic-preserving edits (viewpoint variation, background modification, lighting changes) that break pixel-level correspondence between reference and target while preserving identity. By training with these perturbed references, the model is forced to extract only identity information from the reference and rely on the audio signal for temporal dynamics, rather than copying static content. Detailed prompts are provided in the supplementary material.

\noindent\textbf{Similarity-calibrated sampling.}
Not all perturbed references are equally useful. We filter and organize them using a two-stage criterion: (1)~\emph{Identity filtering}: we compute ArcFace~\cite{deng2019arcface} cosine similarity between the perturbed reference and the original, retaining only pairs with similarity $> \tau_{\text{arc}}$ to ensure identity consistency. (2)~\emph{Perturbation-aware bucketing}: for retained pairs, we compute the structural similarity (SSIM~\cite{wang2004image}) between the perturbed and original images and partition pairs into 5 buckets of width 0.15 spanning $[0.25, 1.0]$. During training, we uniformly sample across buckets, ensuring the model sees a balanced distribution from near-identical to substantially different reference-target pairs. This balanced curriculum prevents the model from relying on the copy shortcut, forcing it to generate genuine audio-driven dynamics.

\begin{table*}[t]
\centering
\caption{Quantitative comparison on GenBench-ShortVideo ($\sim$10\,s) and GenBench-LongVideo ($>$5\,min)~\cite{huang2025liveavatar}. \textbf{Bold}: best. \ul{Underline}: second best. $\uparrow$/$\downarrow$: higher/lower is better. FPS is resolution- and hardware-dependent (720$\times$400, single H800 node) and identical across benchmarks.}
\label{tab:main_comparison}
\label{tab:long_video}
\footnotesize
\tablestyle{5.5pt}{1.2}
\setlength{\aboverulesep}{0.2ex}
\setlength{\belowrulesep}{0.2ex}
\resizebox{0.80\textwidth}{!}{%
\begin{tabular}{l !{\color{gray!60}\vrule width 0.3pt} cccccc !{\color{gray!60}\vrule width 0.3pt} cc}
\toprule
\multirow{2}{*}{\textbf{Method}} & \multicolumn{6}{c!{\color{gray!60}\vrule width 0.3pt}}{\textbf{Standard Metrics}} & \multicolumn{2}{c}{\textbf{Dynamics}} \\
& ASE$\uparrow$ & IQA$\uparrow$ & Sync-C$\uparrow$ & Sync-D$\downarrow$ & Dino-S$\uparrow$ & FPS$\uparrow$ & ExpVar$\uparrow$ & Dyn-Deg$\uparrow$ \\
\shline
\rowcolor{isabelline}
\multicolumn{9}{c}{\textit{\textbf{GenBench-ShortVideo}}} \\ \shline
\rowcolor{isabelline}
\multicolumn{9}{l}{\textit{Non-real-time}} \\ \shline
WanS2V~\cite{gao2025wans2vaudiodrivencinematicvideo} & 3.36 & 4.29 & 5.89 & 9.08 & 0.95 & 0.25 & 1.81 & 0.72 \\
OmniAvatar~\cite{gan2025omniavatar} & \ul{3.53} & 4.49 & 6.77 & \ul{8.22} & 0.95 & 0.16 & \ul{1.95} & \textbf{0.75} \\
Hallo3~\cite{cui2025hallo3} & 3.12 & 3.97 & 4.74 & 10.19 & 0.94 & 0.26 & 1.48 & 0.57 \\
StableAvatar~\cite{tu2025stableavatar} & 3.52 & 4.47 & 3.42 & 11.33 & 0.93 & 0.64 & 1.23 & 0.51 \\
EchoMimic-V2~\cite{meng2025echomimicv2strikingsimplifiedsemibody} & 2.82 & 3.61 & 5.57 & 9.13 & 0.79 & 0.53 & 1.56 & 0.60 \\
\shline
\rowcolor{isabelline}
\multicolumn{9}{l}{\textit{Real-time}} \\ \shline
Ditto~\cite{li2024ditto} & 3.31 & 4.24 & 4.09 & 10.76 & \textbf{0.99} & 21.80 & 0.94 & 0.37 \\
LiveAvatar~\cite{huang2025liveavatar} & 3.44 & 4.51 & 7.03 & 8.30 & \ul{0.96} & \textbf{45.2} & 0.69 & 0.31 \\
SoulX-FlashTalk~\cite{soulxflashtalk2025} & 3.48 & \ul{4.53} & \ul{7.12} & 8.25 & \ul{0.96} & \ul{32.5} & 0.81 & 0.35 \\
\shline
\rowcolor{lightblue}
\textbf{\ours{} (Ours)} & \textbf{3.55} & \textbf{4.58} & \textbf{7.68} & \textbf{7.89} & \ul{0.96} & \textbf{45.2} & \textbf{2.02} & \ul{0.73} \\
\shline
\rowcolor{isabelline}
\multicolumn{9}{c}{\textit{\textbf{GenBench-LongVideo}}} \\ \shline
\rowcolor{isabelline}
\multicolumn{9}{l}{\textit{Non-real-time}} \\ \shline
WanS2V & 2.63 & 3.99 & 6.04 & 9.12 & 0.80 & 0.25 & 1.63 & 0.65 \\
OmniAvatar & 2.36 & 2.86 & \ul{8.00} & \textbf{7.59} & 0.66 & 0.16 & \textbf{2.06} & \textbf{0.71} \\
Hallo3 & 2.65 & 4.04 & 6.18 & 9.29 & 0.83 & 0.26 & 1.36 & 0.53 \\
StableAvatar & 3.00 & 4.66 & 1.97 & 13.57 & 0.94 & 0.64 & 1.07 & 0.44 \\
\shline
\rowcolor{isabelline}
\multicolumn{9}{l}{\textit{Real-time}} \\ \shline
Ditto & 2.90 & 4.48 & 3.98 & 10.57 & \textbf{0.97} & 21.80 & 0.76 & 0.35 \\
LiveAvatar & \ul{3.42} & \ul{4.76} & 7.16 & 8.31 & \textbf{0.97} & \textbf{45.2} & 0.57 & 0.28 \\
SoulX-FlashTalk & 3.40 & 4.72 & 7.08 & 8.18 & \ul{0.96} & \ul{32.5} & 0.67 & 0.32 \\
\shline
\rowcolor{lightblue}
\textbf{\ours{} (Ours)} & \textbf{3.50} & \textbf{4.82} & \textbf{8.05} & \ul{8.02} & \textbf{0.97} & \textbf{45.2} & \ul{1.93} & \ul{0.68} \\
\bottomrule
\end{tabular}
}
\end{table*}

\subsection{Efficient Self-Forcing Training}
\label{sec:efficiency}

Self-forcing training is memory-intensive because the full rollout graph must be retained for backpropagation~\cite{huang2025self}. We introduce two optimizations that substantially reduce memory requirements.

\noindent\textbf{Computation graph pruning.}
We empirically observe that cross-block gradients (propagated through the KV cache) have negligible impact on training outcomes, yet significantly increase memory pressure. We therefore detach the KV cache from the computation graph, removing cross-block gradient dependencies.

\noindent\textbf{Block-wise gradient replay.}
With cross-block gradients eliminated, we adopt a two-phase training procedure. In the \emph{rollout phase}, the entire sequence is generated with gradients disabled, and intermediate results (latent outputs, KV cache entries) are stored in memory. In the \emph{replay phase}, we iterate over blocks one at a time: for each block, we re-enable gradients and run a single forward pass using the stored intermediates, then compute and accumulate the DMD gradient. This reduces peak activation memory by $\sim$$K{\times}$ (where $K$ is the number of blocks), at the cost of $K$ additional forward passes.

\section{Experiments}
\label{sec:experiments}

\subsection{Experimental Setup}

\noindent\textbf{Implementation details.} 
Our approach is built upon the WanS2V~\cite{gao2025wans2vaudiodrivencinematicvideo} (14B). The training process is divided into two stages. For Stage~1, we strictly follow the diffusion-forcing pretraining protocol introduced by LiveAvatar~\cite{huang2025liveavatar}, while our proposed modifications are exclusively applied during Stage~2. During the inference phase, we employ the TPP strategy from~\cite{huang2025liveavatar} to generate videos at a resolution of 720$\times$400. Regarding the optimization hyperparameters, we set the probability $p_{\text{data}}$ to $0.3$. We intentionally concentrate the starting-step probability mass on the early noise levels by assigning $p(t_1) = 0.70$, alongside $p(t_0) = p(t_2) = 0.15$. Furthermore, the reward weights are configured as follows: the synchronization weight $\lambda_{\text{sync}}$ is $1.0$, the expression weight $\lambda_{\text{exp}}$ is $0.5$, and the balancing factor $\alpha$ is $0.1$. We also establish an ArcFace threshold of $\tau_{\text{arc}} = 0.9$. Finally, our model is trained for 4,000 steps on eight NVIDIA H100 GPUs. For the DMD optimization, we use a learning rate of $1 \times 10^{-5}$ for the student and $2 \times 10^{-6}$ for the fake score, with an update frequency ratio of 5.

\noindent\textbf{Datasets.} 
We train our model on the AVSpeech dataset~\cite{AVSpeech}, utilizing the preprocessing pipeline introduced by OmniAvatar~\cite{gan2025omniavatar}. This resulting training corpus consists of 400,000 samples, with all video clips having a duration of more than 10 seconds. For evaluation, we adopt two benchmark datasets from~\cite{huang2025liveavatar}. The first is GenBench-ShortVideo, which includes 100 samples that are approximately 10 seconds long, and the second is GenBench-LongVideo, which comprises 15 samples with durations exceeding 5 minutes.

\noindent\textbf{Evaluation metrics.}
We adopt standard metrics: Q-Align~\cite{qalign} for perceptual quality (IQA) and aesthetic appeal (ASE), Sync-C and Sync-D~\cite{syn} for audio-visual synchronization, DINOv2~\cite{oquab2023dinov2} cosine similarity (Dino-S) for identity consistency. We additionally report \textbf{dynamics metrics}: \textbf{ExpVar} (standard deviation of 3DMM expression coefficients across frames~\cite{zhu2025infp}, measuring expression dynamics) and VBench~\cite{huang2023vbench} \textbf{Dyn-Deg} (dynamic degree, measuring temporal variation diversity).

\noindent\textbf{Baselines.}
We compare with methods spanning two paradigms: (1)~\emph{Non-real-time} multi-step diffusion models: WanS2V~\cite{gao2025wans2vaudiodrivencinematicvideo} (teacher), OmniAvatar~\cite{gan2025omniavatar}, Hallo3~\cite{cui2025hallo3}, StableAvatar~\cite{tu2025stableavatar}, EchoMimic-V2~\cite{meng2025echomimicv2strikingsimplifiedsemibody}; (2)~\emph{Real-time} streaming models: Ditto~\cite{li2024ditto}, LiveAvatar~\cite{huang2025liveavatar}, SoulX-FlashTalk~\cite{soulxflashtalk2025}. All benchmarked at 720$\times$400 on a single H800 node. For methods with strict requirements on reference image aspect ratio or subject positioning (SoulX-FlashTalk and EchoMimic-V2), we use their official open-source inference pipelines without modification to ensure fair comparison. The ``self-forcing baseline'' in ablations denotes standard self-forcing trained for 4,000 steps, identical to \ours{} but without our three proposed components.

\subsection{Main Results}
\begin{figure*}[!t]
  \centering
  \includegraphics[width=0.78\linewidth]{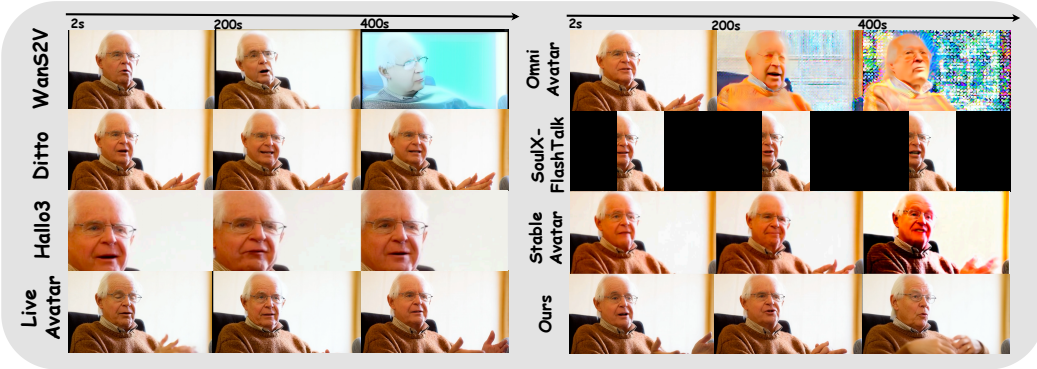}
  \caption{Qualitative comparison on GenBench-LongVideo. Frames sampled at 2\,s, 200\,s, and 400\,s from the same driving audio. Please refer to the supplementary video for dynamics comparison.}
  \label{fig:qualitative}
\end{figure*}

\noindent\textbf{GenBench-ShortVideo.}
Table~\ref{tab:main_comparison} (top) presents results. Real-time self-forcing models (LiveAvatar, SoulX-FlashTalk) achieve strong visual quality but suffer severely on dynamics: LiveAvatar's ExpVar (0.69) is 62\% lower than its teacher WanS2V (1.81), and its Dyn-Deg (0.31) is less than half of OmniAvatar (0.75), confirming dynamic collapse as a systematic issue. \ours{} recovers dynamics to teacher-comparable levels: Dyn-Deg rises from 0.31 to 0.73 (teacher: 0.72), ExpVar from 0.69 to 2.02, and Sync-C from 7.03 to 7.68. That ExpVar slightly exceeds the teacher (2.02 vs.\ 1.81) is consistent with the known tendency of DMD-distilled models to concentrate output distributions~\cite{imageteam2025zimage,luo2025learning}; our reward regularization further steers this concentration toward dynamically-rich modes. Note that Sync-C and ExpVar are both components of our training reward, so gains on these two metrics alone cannot rule out reward hacking. We therefore highlight Dyn-Deg, which is \emph{not} optimized during training yet recovers from 0.31 to 0.73 (matching the teacher's 0.72), as independent evidence of genuine dynamics recovery. Our user study further corroborates this with strong human preference on motion naturalness. Moreover, by stabilizing dynamics throughout training, \ours{} avoids the early quality--dynamics trade-off that forces vanilla self-forcing to stop prematurely, enabling full convergence and thus improved visual quality as a byproduct (ASE/IQA: 3.55/4.58 vs.\ 3.44/4.51). FPS remains 45.2 as our contributions are training-side only.

\noindent\textbf{GenBench-LongVideo.}
Table~\ref{tab:main_comparison} (bottom) shows that the dynamics advantage of \ours{} holds on long-form generation ($>$5\,min). LiveAvatar's ExpVar further drops from 0.69 (short) to 0.57 (long) as static representations accumulate in the KV cache, while \ours{} maintains strong dynamics (ExpVar 1.93) alongside the best visual quality and synchronization.

\noindent\textbf{Qualitative results.}
Figure~\ref{fig:qualitative} shows representative frames from all compared methods. We refer readers to the supplementary video for a more faithful comparison of expression dynamics, which static frames cannot fully convey.

\begin{figure}[t]
  \centering
  \includegraphics[width=0.86\linewidth]{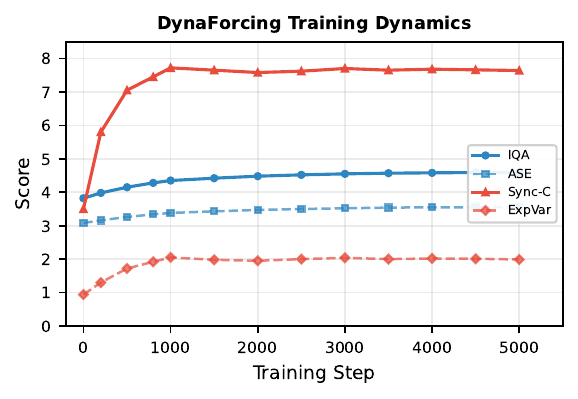}
  \vspace{-0.5cm}
  \caption{Training dynamics of \ours{}. Unlike vanilla self-forcing (Figure~\ref{fig:dynamic_collapse}b), dynamics metrics (Sync-C, ExpVar) stabilize after $\sim$1K steps without degradation, while quality metrics improve steadily.}
  \label{fig:dynaforcing_curve}
  \vspace{-0.1cm}
\end{figure}

\noindent\textbf{Training stability.}
Figure~\ref{fig:dynaforcing_curve} shows the training dynamics of our full \ours{}. Unlike vanilla self-forcing (Figure~\ref{fig:dynamic_collapse}b), where dynamics metrics peak early and then collapse, \ours{}'s Sync-C and ExpVar rise within the first $\sim$1,000 steps and remain stable throughout training. Meanwhile, quality metrics (ASE, IQA) steadily improve and plateau around 4,000 steps. This confirms that our method resolves the quality--dynamics trade-off throughout the entire training process, rather than relying on early stopping at a favorable checkpoint.

\subsection{Ablation Studies}
\label{sec:ablation}

All ablations are evaluated on GenBench-ShortVideo unless otherwise noted.

\noindent\textbf{Diagnostic ablation: isolating rollout feedback.}
To test our hypothesis that unanchored self-conditioning amplifies collapse (\S\ref{sec:dynamic_collapse}), we compare two training paradigms under identical settings (4,000 steps, same DMD loss, no \ours{} components): (i)~\emph{Self-forcing}, where each block's context is built from the student's own outputs; (ii)~\emph{GT-anchored (CausVid-style)}, which follows CausVid~\cite{yin2025slow} and processes all blocks in parallel on noised ground-truth latents with a blockwise causal mask. The core difference between the two paradigms is the conditioning source (student output vs.\ ground-truth).

\begin{table}[t]
\centering
\caption{Diagnostic ablation: isolating rollout feedback. GT-anchored training conditions on ground-truth latents; self-forcing conditions on the student's own outputs.}
\label{tab:diagnostic_rollout}
\small
\tablestyle{5.5pt}{1.2}
\setlength{\aboverulesep}{0.2ex}
\setlength{\belowrulesep}{0.2ex}
\begin{tabular}{@{}l !{\color{gray!60}\vrule width 0.3pt} cccc@{}}
\toprule
\textbf{Configuration} & ASE$\uparrow$ & Sync-C$\uparrow$ & ExpVar$\uparrow$ & Dyn-Deg$\uparrow$ \\
\midrule
Self-forcing & \textbf{3.44} & 5.48 & 0.69 & 0.31 \\
\rowcolor{lightblue}
GT-anchored (CausVid-style) & 3.38 & \textbf{6.85} & \textbf{1.36} & \textbf{0.52} \\
\bottomrule
\end{tabular}
\end{table}

Table~\ref{tab:diagnostic_rollout} reveals a stark contrast. Self-forcing achieves slightly higher visual quality (ASE) but severely suppressed dynamics (Sync-C: 5.48, Dyn-Deg: 0.31), confirming dynamic collapse under unanchored self-conditioning. GT-anchored training, which anchors every block to ground-truth context, recovers substantial dynamics (ExpVar: 0.69$\to$1.36, Dyn-Deg: 0.31$\to$0.52) and lip-sync accuracy (Sync-C: 5.48$\to$6.85). This supports our claim that the conditioning source, not DMD distillation itself, is the primary driver of collapse. However, GT-anchored training alone does not fully close the gap to the teacher (ExpVar still below 1.81), indicating that reverse-KL mode-seeking contributes independently.

\noindent\textbf{Component-wise ablation.}
\begin{table}[t]
\centering
\caption{Component ablation. Each row removes one component from full \ours{}.}
\label{tab:ablation_component}
\small
\tablestyle{5.5pt}{1.2}
\setlength{\aboverulesep}{0.2ex}
\setlength{\belowrulesep}{0.2ex}
\begin{tabular}{@{}l !{\color{gray!60}\vrule width 0.3pt} ccc@{}}
\toprule
\textbf{Configuration} & ASE$\uparrow$ & Sync-C$\uparrow$ & ExpVar$\uparrow$ \\
\midrule
\ours{} w/o Hybrid Forcing & \textbf{3.57} & 6.78 & 1.18 \\
\ours{} w/o Reward Reg. & 3.55 & 7.35 & 1.75 \\
\ours{} w/o Ref. Perturbation & 3.52 & 7.58 & 1.92 \\
\shline
\rowcolor{lightblue}
\textbf{\ours{} (full)} & 3.55 & \textbf{7.68} & \textbf{2.02} \\
\bottomrule
\end{tabular}
\end{table}
Table~\ref{tab:ablation_component} presents component contributions. Removing Hybrid Forcing causes the largest drop in both dynamics (ExpVar 2.02$\to$1.18) and lip-sync (Sync-C 7.68$\to$6.78), confirming data-level anchoring as the most critical component; notably, ASE slightly increases (3.55$\to$3.57), consistent with the quality--dynamics trade-off observed in the $p_{\text{data}}$ sweep. Removing Reward Regularization most degrades Sync-C among the remaining components (7.68$\to$7.35), as it directly optimizes lip-sync via the dynamics reward. Removing Reference Perturbation has the smallest individual effect but still contributes (ExpVar 2.02$\to$1.92). All three together achieve the best overall balance.

\noindent\textbf{Hybrid Forcing probability.}
\begin{table}[t]
\centering
\caption{Effect of Hybrid Forcing probability $p_{\text{data}}$ on full \ours{} (varying $p_{\text{data}}$ only).}
\label{tab:ablation_pdata}
\small
\tablestyle{5.5pt}{1.2}
\setlength{\aboverulesep}{0.2ex}
\setlength{\belowrulesep}{0.2ex}
\begin{tabular}{c !{\color{gray!60}\vrule width 0.3pt} cccc}
\toprule
$p_{\text{data}}$ & ASE$\uparrow$ & Sync-C$\uparrow$ & ExpVar$\uparrow$ & Dino-S$\uparrow$ \\
\midrule
0.1 & \textbf{3.56} & 7.19 & 1.42 & \textbf{0.96} \\
\rowcolor{lightblue}
\textbf{0.3} & 3.55 & \textbf{7.68} & 2.02 & \textbf{0.96} \\
0.5 & 3.53 & 7.64 & 2.03 & 0.95 \\
1.0 & 3.42 & 7.62 & \textbf{2.10} & 0.93 \\
\bottomrule
\end{tabular}
\end{table}
Table~\ref{tab:ablation_pdata} sweeps the mixing probability. $p_{\text{data}} = 0.3$ achieves the best overall balance. Reducing $p_{\text{data}}$ to 0.1 yields slightly higher visual quality (ASE 3.56) but substantially degrades lip-sync (Sync-C 7.68$\to$7.19) and dynamics (ExpVar 2.02$\to$1.42), as the model receives insufficient ground-truth anchoring. Increasing $p_{\text{data}}$ beyond 0.3 continues to improve ExpVar (reaching 2.10 at $p_{\text{data}} = 1.0$) thanks to stronger ground-truth anchoring, but at the cost of significant train-test mismatch: the model gets less self-forcing practice, degrading visual quality (ASE 3.55$\to$3.42) and identity consistency (Dino-S 0.96$\to$0.93) at inference. $p_{\text{data}} = 0.3$ thus strikes the best balance between dynamics recovery and generation quality.

\begin{table}[t]
\centering
\vspace{-2mm}
\caption{Training efficiency: full \ours{} with and without our efficiency optimizations (\S\ref{sec:efficiency}), both trained for 4,000 steps.}
\label{tab:efficiency}
\small
\tablestyle{5.5pt}{1.2}
\setlength{\aboverulesep}{0.2ex}
\setlength{\belowrulesep}{0.2ex}
\begin{tabular}{@{}l !{\color{gray!60}\vrule width 0.3pt} cc !{\color{gray!60}\vrule width 0.3pt} ccc@{}}
\toprule
 & GPUs & GPU$\cdot$h & ASE$\uparrow$ & Sync-C$\uparrow$ & ExpVar$\uparrow$ \\
\midrule
w/o eff.\ optim. & 128$\times$H100 & 7{,}111 & \textbf{3.56} & 7.66 & \textbf{2.04} \\
\rowcolor{lightblue}
\textbf{w.\ eff.\ optim.} & \textbf{8$\times$H100} & \textbf{667} & 3.55 & \textbf{7.68} & 2.02 \\
\bottomrule
\end{tabular}
\vspace{-3mm}
\end{table}

\noindent\textbf{Training efficiency.}
Table~\ref{tab:efficiency} compares full \ours{} trained with and without our efficiency optimizations. Graph pruning combined with block-wise gradient replay reduces the GPU requirement from 128 to 8 H100s, and total GPU-hours from 7,111 to 667 ($\sim$10.7$\times$ reduction) despite a $\sim$1.5$\times$ per-step wall-clock overhead from replay. Quality metrics remain essentially unchanged (ASE differs by 0.01, Sync-C by 0.02, ExpVar by 0.02), confirming negligible degradation from the approximations.

\noindent\textbf{User study.}
A pairwise study with 25 participants finds \ours{} consistently preferred over both real-time baselines, most strongly on motion naturalness (80.6\% vs.\ LiveAvatar); since raters have no access to our training rewards, this independently supports genuine dynamics recovery (see supplementary material for the full protocol and results).

\section{Conclusion}
\label{sec:conclusion}

In the context of audio-driven avatar generation, we have formalized and systematically characterized \emph{dynamic collapse}, a failure mode of self-forcing distillation where the reverse KL objective and autoregressive feedback jointly drive the student toward near-static outputs, suppressing the temporal dynamics essential for faithful talking-head synthesis. To address this, we proposed \textbf{\ours{}}, which intervenes at the data, loss, and conditioning levels of training, together with computation graph pruning and gradient replay that reduce GPU requirements by $\sim$10$\times$.

\noindent\textbf{Limitations.}
Our dynamics reward relies on pretrained SyncNet and 3DMM models, whose biases may influence training. Our analysis centers on avatar generation; while we expect the data-level component to transfer, systematic validation across other tasks remains future work. The $\sim$1.5$\times$ overhead from gradient replay may also be further reducible.

\begin{acks}
This work was supported in part by the grant from National Natural Science Foundation of China (No.~62406264), and the grant from the Anhui Natural Science Foundation (No.~2508085ZD006).
\end{acks}

\bibliographystyle{ACM-Reference-Format}
\clearpage
\bibliography{references}

\clearpage
\appendix
\section{Overview of Supplementary Material}
\label{app:overview}

This supplementary document provides additional details and experiments to support the main paper. The content is organized as follows:

\begin{itemize}
\item \textbf{Section~\ref{app:ref_perturb_prompts}: Reference Perturbation Prompts.} Full prompt templates used for generating perturbed reference images via Qwen-Image-Edit.

\item \textbf{Section~\ref{app:collapse_setup}: Reproduction Setup.} Training configurations used to reproduce the dynamic collapse curves in Figure~2 (main paper).

\item \textbf{Section~\ref{app:additional_ablations}: Additional Ablation Studies.} Supplementary ablations on reward strength $\alpha$, starting-step distribution, and reward components.

\item \textbf{Section~\ref{app:perturb_strategy}: Perturbation Strategy Comparison.} Comparison of Qwen-Image-Edit against traditional data augmentation for reference perturbation.

\item \textbf{Section~\ref{app:reward_hacking}: Reward-Free Dynamics Evaluation.} Landmark-based motion statistics that verify dynamics recovery is genuine rather than reward hacking.

\item \textbf{Section~\ref{app:user_study}: User Study Details.} Protocol, setup, and full per-dimension results of the perceptual user study.

\item \textbf{Section~\ref{app:ethics}: Ethical Considerations.} Discussion of potential risks and responsible deployment practices.
\end{itemize}

\section{Reference Perturbation Prompts}
\label{app:ref_perturb_prompts}

We use Qwen-Image-Edit~\cite{wu2025qwenimagetechnicalreport} to generate perturbed reference images that preserve identity but vary in other visual attributes. For each training identity, we apply four prompt categories as shown in Table~\ref{tab:ref_perturb_prompts}. For each source identity, we generate 8--12 perturbed variants spanning all categories. All generated images are filtered by ArcFace~\cite{deng2019arcface} cosine similarity $> 0.9$ to ensure identity consistency before entering the similarity-calibrated bucketing pipeline described in the main paper (Section 4.3).

\begin{table}[t]
\centering
\caption{Full prompts for reference image perturbation via Qwen-Image-Edit. Each category targets a specific visual attribute while explicitly constraining identity-related features to remain unchanged. \texttt{\{$\cdot$\}} denotes a slot randomly sampled from the provided options.}
\label{tab:ref_perturb_prompts}
\small
\renewcommand{\arraystretch}{1.25}
\begin{tabular}{@{}p{0.97\columnwidth}@{}}
\toprule
\textbf{Category 1: Background Replacement} \\
\midrule
\texttt{Change the background to \{a studio with plain gray backdrop / an outdoor park scene / a dimly lit office\}; keep the person's face, hairstyle, expression, and clothing unchanged.} \\
\midrule
\textbf{Category 2: Lighting Variation} \\
\midrule
\texttt{Adjust the lighting to \{warm golden-hour sunlight from the left / cool fluorescent overhead lighting / dramatic side lighting with strong shadows\}; keep the person's appearance unchanged.} \\
\midrule
\textbf{Category 3: Viewpoint Shift} \\
\midrule
\texttt{Slightly rotate the person's head to \{face 15 degrees to the left / tilt slightly upward / turn to a three-quarter view\}; keep the person's identity, hairstyle, and clothing unchanged.} \\
\midrule
\textbf{Category 4: Accessory Modification} \\
\midrule
\texttt{\{Add a pair of thin-framed glasses / Remove earrings / Add a dark scarf around the neck\}; keep the person's face, hairstyle, and expression unchanged.} \\
\bottomrule
\end{tabular}
\end{table}

\section{Reproduction Setup for Figure~2 (main paper)}
\label{app:collapse_setup}

All training curves in Figure~2 (main paper) are produced by reproducing the distillation stage of existing methods using their official codebases.
For the text-to-video experiment (Figure~2 (main paper)a), we use the official Self-Forcing~\cite{huang2025self} code with its default configuration and Stage-1 pretrained weights, then run self-forcing (Stage~2) for 8K steps.
For the audio-driven avatar experiment (Figure~2 (main paper)b), we similarly use the official LiveAvatar~\cite{huang2025liveavatar} code with its released Stage-1 checkpoint and default hyperparameters, then run self-forcing distillation for 8K steps.
For the CausVid comparison (Figure~2 (main paper)c), we use the official CausVid~\cite{yin2025slow} codebase with its released ODE initialization checkpoint and default DMD training configuration, training for 8K steps on the same evaluation prompts.
In all cases, hyperparameters strictly follow the respective original papers; we only extend the training horizon to observe the long-term dynamics trajectory. VBench~\cite{huang2023vbench} dynamic degree is computed at intermediate checkpoints for (a) and (c), while Sync-C, ExpVar, IQA, and ASE are evaluated on a held-out set for (b).

\section{Additional Ablation Studies}
\label{app:additional_ablations}

All supplementary ablations are evaluated on GenBench-ShortVideo with the full \ours{} model, varying only the parameter under study.

\noindent\textbf{Reward strength $\alpha$.}
Table~\ref{tab:alpha_sweep} sweeps the reward strength $\alpha$ in Eq.~7 (main paper). At $\alpha = 0$ (no reward weighting), dynamics degrade to the level of the ``w/o Reward Reg.''\ ablation in the main paper (Table~3 (main paper)). Increasing $\alpha$ to 0.1 yields the best quality--dynamics balance. Beyond 0.2, the exponential weighting over-amplifies high-reward samples, causing visual quality to drop (ASE: 3.55$\to$3.48 at $\alpha = 0.5$) without further dynamics gains (ExpVar saturates around 2.06--2.08). We select $\alpha = 0.1$ for all experiments.

\begin{table}[h]
\centering
\caption{Effect of reward strength $\alpha$ on full \ours{}.}
\label{tab:alpha_sweep}
\small
\tablestyle{5.5pt}{1.2}
\setlength{\aboverulesep}{0.2ex}
\setlength{\belowrulesep}{0.2ex}
\begin{tabular}{c !{\color{gray!60}\vrule width 0.3pt} ccc}
\toprule
$\alpha$ & ASE$\uparrow$ & Sync-C$\uparrow$ & ExpVar$\uparrow$ \\
\midrule
0.0 & \textbf{3.55} & 7.35 & 1.75 \\
0.05 & \textbf{3.55} & 7.52 & 1.88 \\
\rowcolor{lightblue}
\textbf{0.1} & \textbf{3.55} & 7.68 & 2.02 \\
0.2 & 3.53 & \textbf{7.70} & \textbf{2.08} \\
0.5 & 3.48 & 7.65 & 2.06 \\
\bottomrule
\end{tabular}
\end{table}

\noindent\textbf{Starting-step probability distribution.}
Table~\ref{tab:startstep_sweep} compares different distributions for sampling $t_{\text{start}}$ in data-anchored mode (Eq.~5 (main paper)). ``Conc.\ $t_k$'' concentrates 70\% mass on step $t_k$ with 15\% on each of the other two. Concentrating on $t_0$ (highest SNR) wastes training on near-trivial denoising, while concentrating on $t_2$ (lowest SNR) loses too much ground-truth structure. Concentrating on $t_1$ achieves the best balance, and outperforms the uniform baseline.

\begin{table}[h]
\centering
\caption{Effect of starting-step probability distribution.}
\label{tab:startstep_sweep}
\small
\tablestyle{5.5pt}{1.2}
\setlength{\aboverulesep}{0.2ex}
\setlength{\belowrulesep}{0.2ex}
\begin{tabular}{l !{\color{gray!60}\vrule width 0.3pt} ccc}
\toprule
\textbf{Distribution} & ASE$\uparrow$ & Sync-C$\uparrow$ & ExpVar$\uparrow$ \\
\midrule
Uniform & 3.54 & 7.51 & 1.86 \\
Conc.\ $t_0$ & \textbf{3.55} & 7.42 & 1.78 \\
\rowcolor{lightblue}
\textbf{Conc.\ $t_1$} & \textbf{3.55} & \textbf{7.68} & \textbf{2.02} \\
Conc.\ $t_2$ & 3.52 & 7.55 & 1.91 \\
\bottomrule
\end{tabular}
\end{table}

\noindent\textbf{Reward component analysis.}
Table~\ref{tab:reward_components} isolates the contribution of each reward component in Eq.~8 (main paper). Using only Sync-C reward ($\lambda_{\text{exp}} = 0$) primarily improves lip-sync but provides limited expression dynamics recovery. Using only ExpVar reward ($\lambda_{\text{sync}} = 0$) improves expression variance but is less effective at lip-sync. Combining both yields the best overall performance, confirming their complementary roles.

\begin{table}[h]
\centering
\caption{Ablation on reward components in $R_{\text{dyn}}$.}
\label{tab:reward_components}
\small
\tablestyle{5.5pt}{1.2}
\setlength{\aboverulesep}{0.2ex}
\setlength{\belowrulesep}{0.2ex}
\begin{tabular}{l !{\color{gray!60}\vrule width 0.3pt} ccc}
\toprule
\textbf{Reward} & ASE$\uparrow$ & Sync-C$\uparrow$ & ExpVar$\uparrow$ \\
\midrule
None ($\alpha = 0$) & \textbf{3.55} & 7.35 & 1.75 \\
Sync-C only & \textbf{3.55} & 7.62 & 1.85 \\
ExpVar only & 3.54 & 7.48 & 1.96 \\
\rowcolor{lightblue}
\textbf{Both (full)} & \textbf{3.55} & \textbf{7.68} & \textbf{2.02} \\
\bottomrule
\end{tabular}
\end{table}

\section{Perturbation Strategy Comparison}
\label{app:perturb_strategy}

Reference Perturbation (main paper, Section~4.3) uses Qwen-Image-Edit to generate semantically edited references. A natural question is whether this comparatively heavy pipeline is necessary, or whether conventional data augmentation would suffice. We therefore replace Qwen-Image-Edit with traditional augmentations while keeping the ArcFace filtering and similarity-calibrated bucketing unchanged. For each reference we randomly apply one of: (1)~patch dropout ($30\times30$ patches, drop rate ${\sim}\mathcal{U}(0, 0.5)$), (2)~color jitter (strength ${\sim}\mathcal{U}(0, 0.8)$), or (3)~perspective rotation (${\sim}\mathcal{U}(0^\circ, 30^\circ)$).

\begin{table}[h]
\centering
\caption{Perturbation strategy comparison on GenBench-ShortVideo.}
\label{tab:perturb_strategy}
\small
\tablestyle{5.5pt}{1.2}
\setlength{\aboverulesep}{0.2ex}
\setlength{\belowrulesep}{0.2ex}
\begin{tabular}{l !{\color{gray!60}\vrule width 0.3pt} ccc}
\toprule
\textbf{Perturbation} & ASE$\uparrow$ & Sync-C$\uparrow$ & ExpVar$\uparrow$ \\
\midrule
None & 3.52 & 7.58 & 1.92 \\
Traditional Aug. & 3.48 & \textbf{7.70} & \textbf{2.03} \\
\rowcolor{lightblue}
\textbf{Qwen-Image-Edit (Ours)} & \textbf{3.55} & 7.68 & 2.02 \\
\bottomrule
\end{tabular}
\end{table}

Table~\ref{tab:perturb_strategy} shows two things. First, the dynamics benefit of reference perturbation is largely \emph{robust to the specific strategy}: traditional augmentation recovers Sync-C and ExpVar to essentially the same level as semantic editing, indicating that what matters is breaking the pixel-level correspondence between reference and target rather than the particular mechanism used. Second, traditional augmentation slightly degrades visual quality (ASE: 3.55$\to$3.48), plausibly because unrealistic corruptions shift references off the natural image distribution, whereas semantic edits remain in-distribution. We also note that Qwen-Image-Edit is applied \emph{offline} to the 32K training samples used for distillation, so its cost is negligible relative to training.

\section{Reward-Free Dynamics Evaluation}
\label{app:reward_hacking}

Sync-C and ExpVar are used as training rewards, so improvements on these two metrics alone cannot exclude reward hacking. The main paper therefore highlights Dyn-Deg, which is not optimized during training. Here we supplement two further reward-free statistics computed from 2D facial landmarks extracted with FAN~\cite{bulat2017far}, using the 68-point mouth region: \textbf{Lip-Vel}, the mean first-order displacement $\lVert L_{t} - L_{t-1} \rVert$ of lip landmarks, which measures motion magnitude; and \textbf{Lip-Jit}, the mean second-order difference $\lVert L_{t+1} - 2L_{t} + L_{t-1} \rVert$, which measures smoothness and is sensitive to jitter. Higher Lip-Vel indicates more lip motion; \emph{lower} Lip-Jit indicates smoother motion.

\begin{table}[h]
\centering
\caption{Reward-free dynamics evaluation. Sync-C and ExpVar are training rewards; Dyn-Deg, Lip-Vel and Lip-Jit are not optimized.}
\label{tab:reward_free}
\scriptsize
\tablestyle{2pt}{1.1}
\setlength{\aboverulesep}{0.2ex}
\setlength{\belowrulesep}{0.2ex}
\begin{tabular}{l !{\color{gray!60}\vrule width 0.3pt} cc !{\color{gray!60}\vrule width 0.3pt} ccc}
\toprule
\multirow{2}{*}{\textbf{Method}} & \multicolumn{2}{c!{\color{gray!60}\vrule width 0.3pt}}{\textit{Reward metrics}} & \multicolumn{3}{c}{\textit{Reward-free}} \\
& Sync-C$\uparrow$ & ExpVar$\uparrow$ & Dyn-Deg$\uparrow$ & Lip-Vel$\uparrow$ & Lip-Jit$\downarrow$ \\
\midrule
LiveAvatar & 7.03 & 0.69 & 0.31 & 3.50 & 8.64 \\
\ours{} w/o Reward Reg. & 7.35 & 1.75 & 0.55 & 3.41 & \textbf{4.23} \\
\rowcolor{lightblue}
\textbf{\ours{} (full)} & \textbf{7.68} & \textbf{2.02} & \textbf{0.73} & \textbf{3.78} & 4.26 \\
\bottomrule
\end{tabular}
\end{table}

Table~\ref{tab:reward_free} shows that adding Reward Regularization improves not only the reward metrics but also the reward-free ones: Dyn-Deg rises from 0.55 to 0.73 and Lip-Vel from 3.41 to 3.78. Critically, Lip-Jit is essentially unchanged (4.23 vs.\ 4.26), meaning the additional motion is not accompanied by increased frame-to-frame jitter. This addresses the concern that \ours{}'s ExpVar exceeding the teacher's (2.02 vs.\ 1.81) might reflect over-expression or unnatural trembling rather than richer dynamics: if the reward were being hacked by injecting high-frequency noise, Lip-Jit would rise sharply. Both baselines and \ours{} also improve markedly over LiveAvatar's Lip-Jit of 8.64, whose high value reflects the unstable residual motion left after collapse. We emphasize that Lip-Vel and Lip-Jit are statistics we define here rather than established benchmarks; they are intended as corroborating evidence alongside Dyn-Deg and the user study (Section~\ref{app:user_study}).

\section{User Study Details}
\label{app:user_study}

We conduct a pairwise user study comparing \ours{} against the two strongest competing real-time methods, LiveAvatar~\cite{huang2025liveavatar} and SoulX-FlashTalk~\cite{soulxflashtalk2025}. We recruit 25 participants (14 male, 11 female, aged 22--35) who are na\"ive to the purpose of the study. Each trial presents two videos side-by-side (one from \ours{}, one from a baseline) with synchronized audio. Left/right placement and trial order are randomized. Participants select their preference on four dimensions: visual quality, lip-sync accuracy, motion naturalness, and overall preference. Each participant evaluates 40 trials (20 samples $\times$ 2 baselines), drawn from GenBench-ShortVideo. Paired $t$-tests confirm statistical significance ($p < 0.01$) across all four dimensions.

\begin{table}[h]
\centering
\caption{User study: preference rate~(\%) for \ours{} over the two strongest competing real-time methods.}
\label{tab:user_study}
\small
\tablestyle{5.5pt}{1.2}
\setlength{\aboverulesep}{0.2ex}
\setlength{\belowrulesep}{0.2ex}
\begin{tabular}{l !{\color{gray!60}\vrule width 0.3pt} cccc}
\toprule
\textbf{vs.} & Visual & Lip-Sync & Motion & Overall \\
\midrule
\rowcolor{lightblue}
LiveAvatar & 58.4 & 71.2 & 80.6 & 72.8 \\
\rowcolor{lightblue}
SoulX-FlashTalk & 55.2 & 64.8 & 74.2 & 66.4 \\
\bottomrule
\end{tabular}
\end{table}

Table~\ref{tab:user_study} shows that \ours{} is consistently preferred over both baselines on every dimension. The margin is largest on motion naturalness (80.6\% and 74.2\%), which is the dimension our method directly targets, and smallest on visual quality (58.4\% and 55.2\%), consistent with the narrow ASE/IQA gaps reported in the main paper. Since raters have no access to the training rewards, these preferences corroborate the reward-free metrics of Section~\ref{app:reward_hacking} in indicating that the recovered dynamics are genuine rather than an artifact of reward optimization.

\section{Ethical Considerations}
\label{app:ethics}

Audio-driven avatar generation carries the risk of misuse for creating unauthorized impersonations. We intend our method exclusively for legitimate applications (virtual assistants, telepresence, authorized content creation) and advocate integrating invisible watermarking~\cite{fernandez2023stable} and deepfake detection~\cite{wang2020cnn} into deployment pipelines. We train on AVSpeech~\cite{AVSpeech}, a publicly available dataset with no personally identifiable metadata. Our efficient training strategy reduces compute from $\sim$7{,}111 to $\sim$667 H100 GPU-hours, lowering energy consumption accordingly.

\end{document}